\documentclass[sigconf]{acmart} 

\usepackage{amsmath,amssymb}
\usepackage{natbib}
\usepackage{graphicx}
\usepackage{textcomp}
\usepackage{enumitem}
\usepackage{hyperref}
\usepackage{balance} 
\usepackage{color,soul}
\usepackage{xcolor}
\usepackage{comment}
\usepackage{hhline}
\usepackage{caption}
\usepackage{subcaption}
\usepackage{tikz}
\usetikzlibrary{positioning}
\usepackage{pgfplots}
\usepackage{pdflscape}
\usepackage{pgfplotstable}
\pgfplotsset{compat=1.7}
\usepgfplotslibrary{groupplots}
\usepackage{multirow}
\usepackage{algorithm}

\usepackage{comment}
\usepackage{colortbl}

\def\BibTeX{{\rm B\kern-.05em{\sc i\kern-.025em b}\kern-.08em
    T\kern-.1667em\lower.7ex\hbox{E}\kern-.125emX}}

\makeatletter
\def\addlegendimage{\pgfplots@addlegendimage}
\makeatother

\renewcommand\footnotetextcopyrightpermission[1]{} 

\begin{document}

\title[Exploring Robot Error Detection Using Instrumented Bystander Reactions]{``Why the face?'': Exploring Robot Error Detection \\ Using Instrumented Bystander Reactions}


\author{Maria Teresa Parreira}
\email{mb2554@cornell.edu}
\affiliation{
    \institution{Cornell University, Cornell Tech}
    \state{New York}
    \country{USA}
}
\author{Ruidong Zhang}
\affiliation{
    \institution{Cornell University}
    \state{New York}
    \country{USA}
}
\author{Sukruth Gowdru Lingaraju}
\affiliation{
    \institution{Cornell Tech}
    \state{New York}
    \country{USA}
}
\author{Alexandra Bremers}
\affiliation{
    \institution{Cornell Tech}
    \state{New York}
    \country{USA}
}
\author{Xuanyu Fang}
\affiliation{
    \institution{Cornell Tech}
    \state{New York}
    \country{USA}
}
\author{Adolfo Ramirez-Aristizabal}
\affiliation{
    \institution{Accenture Labs}
    \state{SF}
    \country{USA}
}
\author{Manaswi Saha}
\affiliation{
    \institution{Accenture Labs}
    \state{SF}
    \country{USA}
}
\author{Michael Kuniavsky}
\affiliation{
    \institution{Accenture Labs}
    \state{SF}
    \country{USA}
}
\author{Cheng Zhang}
\affiliation{
    \institution{Cornell University}
    \state{New York}
    \country{USA}
}
\author{Wendy Ju}
\affiliation{
    \institution{Cornell University, Cornell Tech}
    \state{New York}
    \country{USA}
}

\renewcommand{\shortauthors}{Parreira et al.}

\begin{abstract}
How do humans recognize and rectify social missteps? We achieve social competence by looking around at our peers, decoding subtle cues from bystanders — a raised eyebrow, a laugh — to evaluate the environment and our actions. Robots, however, struggle to perceive and make use of these nuanced reactions. By employing a novel neck-mounted device that records facial expressions from the chin region, we explore the potential of previously untapped data to capture and interpret human responses to robot error. First, we develop NeckNet-18, a 3D facial reconstruction model to map the reactions captured through the chin camera onto facial points and head motion. We then use these facial responses to develop a robot error detection model which outperforms standard methodologies such as using OpenFace or video data, generalizing well especially for within-participant data. Through this work, we argue for expanding human-in-the-loop robot sensing, fostering more seamless integration of robots into diverse human environments, pushing the boundaries of social cue detection and opening new avenues for adaptable robotics.
\end{abstract}


\maketitle
\pagestyle{plain}

\section{Introduction}
\label{sec:introduction}




\begin{figure}
    \centering
    \includegraphics[width =1\linewidth]{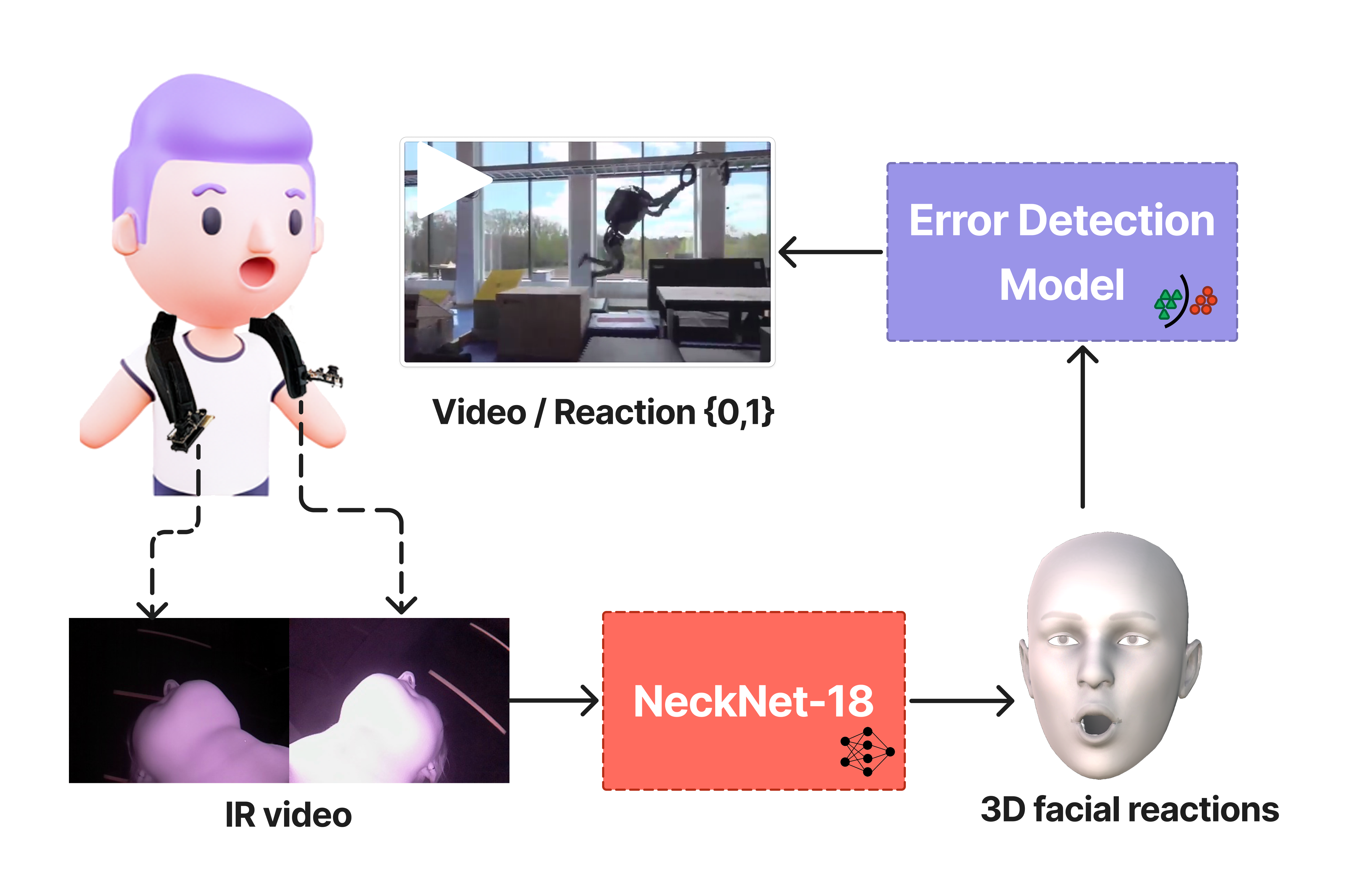}
    \caption{User study scheme. Participants wearing NeckFace \cite{neckface} watch videos where a scenario of human or robot error is shown, eliciting a reaction. The IR camera image is converted into 3D facial points and head rotation data through a customized NeckNet model. This data is then used to train error detection models which map human reactions to the scenario displayed.}
    \label{fig:schema}
\end{figure}

Humans navigate social environments by perceiving and interpreting cues from those around them. \textit{Social competence} stems partly from this ability to read non-verbal signals such as facial expressions, body language, and subtle gestures. These environmental and peer-based inputs form a feedback loop that informs social behavior and facilitates integration into diverse social contexts \cite{junge2020socialcompetence}.

The concept of \textit{human-in-the-loop sensing} in robotics leverages this innate human capability to enhance the social competence and efficiency of robotic systems. This approach posits that human reactions can serve as a rich, real-time source of information for robots, guiding their behavior and functionality. For example, methods for Reinforcement Learning from Human Feedback (RLHF) have leveraged human social cues and emotions as feedback input \cite{jin2020rlhf}. Recent works \cite{stiber2023erroraware,spitale2024errhri2024challengemultimodal, parreira2024badidea} have started exploring the potential of human reactions in the context of human-robot interaction (HRI) for robot error detection. These studies make use of human social cues, captured through various means, as indicators of robot performance and social appropriateness.

Traditional implementations of this concept often rely on stationary cameras or extensive sensor arrays to capture human reactions. While groundbreaking, these methods can be resource-intensive, computationally demanding, and may struggle to adapt to dynamic, real-world environments where both humans and robots are in constant motion. To address these limitations, we employ NeckFace \cite{neckface}, a novel, neck-mounted device that records facial expressions from the chin region and maps them onto 3D facial points and head rotation. This lightweight, bi-camera system offers an alternative perspective on human reactions, potentially capturing nuanced expressions that may be overlooked by traditional methods. Its wearable nature allows for data collection in dynamic, real-world scenarios, providing a more efficient and adaptable solution for understanding human responses to robot behavior.

To test and benchmark the efficiency of NeckFace against other conventional approaches, we conducted a comparative user study. Participants' reactions to video stimuli displaying human and robot error were recorded simultaneously using NeckFace and conventional RGB cameras. Following this stage, we 1) developed and tested a model to map NeckFace's IR camera data onto 3D facial expressions and head motion data, NeckNet-18; 2) used this reaction data to develop error detection models that perform well for unseen participants but also generalize to a single participant with minimal data for training; and 3) compared these models with conventionally used methods, such as OpenFace features \cite{openface, stiber2023erroraware,spitale2024errhri2024challengemultimodal} or RGB camera data \cite{bremers2023bystander, parreira2024study}, finding that it outperforms these methods.

This research not only contributes to the development of more dynamic and context-aware robotic systems but also advances our understanding of how human social competence can be effectively leveraged in HRI. By exploring alternative data sources and expanding on human-in-the-loop sensing methodologies, we aim to pave the way for more socially competent and efficient robotic systems that can seamlessly integrate into human environments.



\section{Related Work}
\label{sec:relwork}

\begin{description}[align=left, leftmargin=0em, labelsep=0.2em, font=\textbf, itemsep=0em,parsep=0.3em]

\item[Social cues for error detection: ]

Human-in-the-loop robotic systems \cite{slade2024humaninloop} leverage human input to enhance robots' efficiency, yielding positive outcomes in HRI \cite{diaz2023humanloop}. Following this concept, harnessing human social cues can equip robots with better social and operational competence. For example, a substantial body of literature has begun exploring social cues in response to robot error, revealing that human reactions are multimodal and diverse \cite{parreira2024badidea,candon2024react}, including body and head motion \cite{kontogiorgos2021systematic,2017trung,giuliani2015systematic}, gaze \cite{cahya2019socialerror}, and facial expressions \cite{kontogiorgos2021systematic,hwang2014reward,2020stiber,2010aronson}. The detection of robot error from social cues has been investigated through various machine learning approaches. \citet{stiber2023erroraware} proposed a framework for ``error aware" HRI, employing a deep neural network, building on previous work \cite{stiber2022effective}. Other studies \cite{spitale2024errhri2024challengemultimodal, parreira2024badidea} utilize recurrent neural networks (RNNs) to preserve time-dependencies. Random Forest classifiers have also demonstrated efficiency as linear methods for error detection \cite{loureiro2023context}. Recently, combinations of Minirocket \cite{minirocket} classifiers have achieved high performance in error and confusion detection \cite{wachowiak2024eyetrack, wachowiak2024errhri}. To bypass the need for feature extraction, some researchers have employed convolutional neural networks \cite{bremers2023bystander, parreira2024study} to classify human reactions directly from video data. Recent work by \citet{ravishankar2024zeroshot} explores unsupervised learning of robot error by detecting abnormalities in human-robot interactions from human-human interaction data. These works explore visual, audio and semantic data modalities. Given the novelty of the field, however, there remains untapped potential for exploring novel data modalities and processing methodologies for social cue classification.

\item[Beyond standard methods for data collection: ]

Capturing facial expressions as humans watch robots can allow the robot to infer human perceptions of robot performance and social appropriateness.
High-accuracy facial expression tracking traditionally involves active motion capture instrumentation on the face, either using visual markers \cite{10.1145/2822013.2822042} or other sensors such as electromyography (EMG) \cite{gruebler2010measurement} or capacitive sensors \cite{rantanen2013capacitive}. These methods usually require extensive setups and calibration and can cause physical and social discomfort. Relieving users from heavy on-face instrumentation, frontal-camera-based vision solutions have been getting recent attention \cite{lugaresi2019mediapipe, baltruvsaitis2016openface}. These solutions usually involve placing a camera in front of the user's face to directly record and analyze their facial expressions. While they provide a less intrusive experience, they are not mobile, as users have to appear in front of the camera within reasonable angle and distance. To address these challenges, recent wearable solutions look more into less intrusive sensing solutions while trying to maintain tracking performance \cite{neckface, bioface3d, eario, eyeecho}.

Among these, NeckFace \cite{neckface} is a wearable facial expression tracking system that can estimate 3-dimensional facial expressions continuously. It performs comparably to traditional video-based systems, without the need for a frontal camera. Unlike other mounted devices, NeckFace is robust to users' head rotations, allowing for greater flexibility in use. It features a 3D-printed neckband with two infrared (IR) cameras positioned on either ends pointing upwards toward the wearer's neck, chin and cheeks (\autoref{fig:schema}). Two 850 nm IR LEDs are placed beside each camera to provide independent lighting to minimize the impact of unstable environmental lighting. NeckFace uses a convolutional neural network (ResNet-34 \cite{he2016deep}) to convert the captured images into 3D facial expressions represented by 52 Blendshapes. NeckFace has been demonstrated to have reliable tracking performance across different scenarios such as sitting, walking and remounting. In light of this, we chose NeckFace as our facial reaction reconstruction device.

To the best of our knowledge, this is the first user study pursuing the use of a neck-mounted device to capture human social cues in HRI, investigating and benchmarking its potential for error detection.

\end{description}

\section{Study Design}\label{sec:methods}

In this section, we detail the NeckFace system implementation, user study design and methodology for data collection. To collect human reactions to observed errors, we followed a procedure adapted from \citet{bremers2023bystander} for in-lab data collection, where participants observed a set of videos where error occurs. The elicitation materials and code used are shared in the project repository \footnote{\url{http://irl.tech.cornell.edu/badrobots-feat-neckface/}}.

\begin{figure}
    \centering
    \includegraphics[width=1\linewidth]{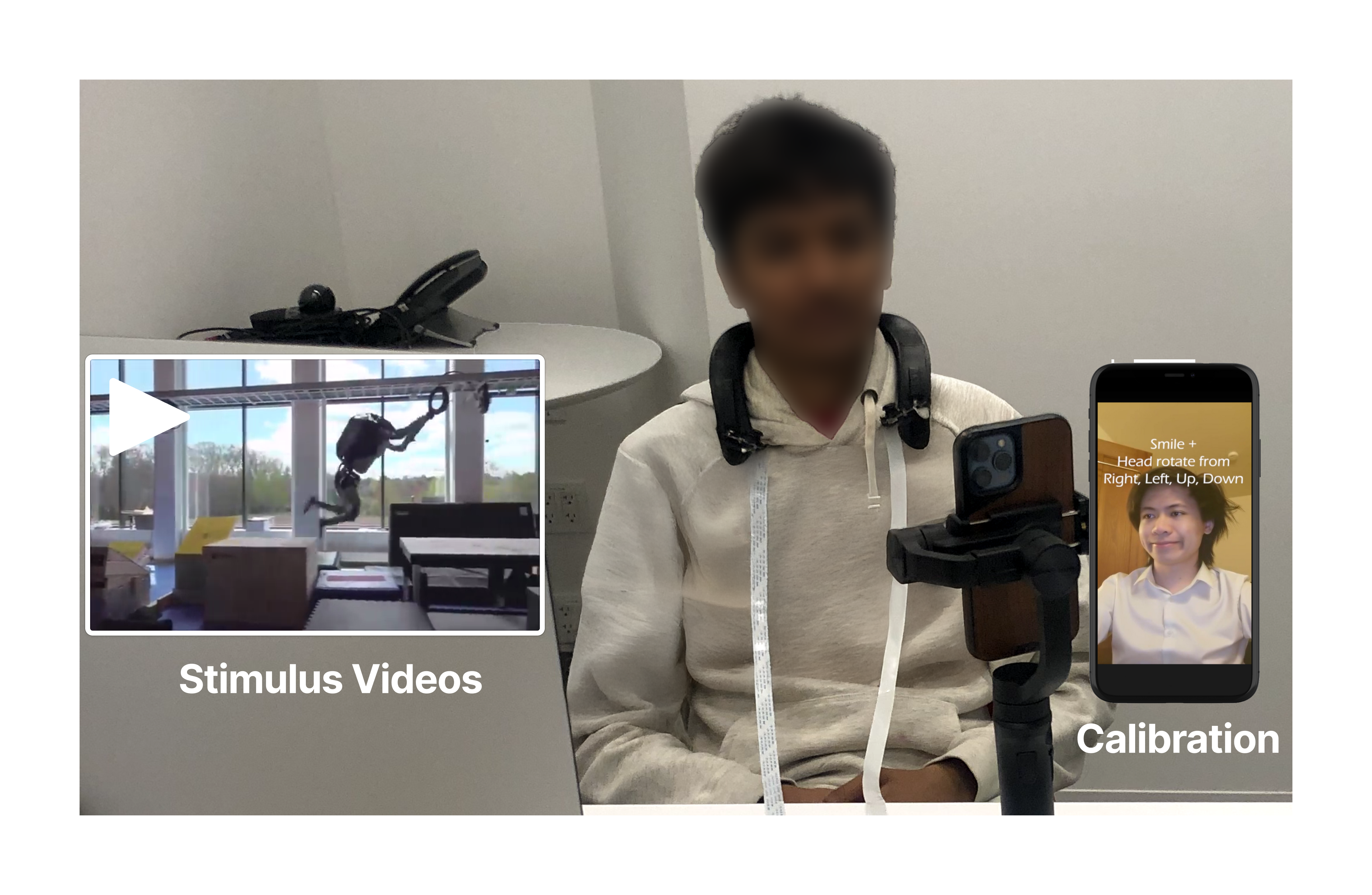}
    \caption{Study setup. In the \textit{Calibration round}, the participant, wearing NeckFace \cite{neckface}, copies the movements seen on a video on an Iphone 11. The \textit{Stimulus round} consists of a series of 30 videos played on a screen, while a webcam and NeckFace record facial reactions.}
    \label{fig:setup}
\end{figure}

\subsection{Study Protocol}
\label{subsec:protocol}

The study set up is shown in \autoref{fig:setup}.
Participants were recruited through flyers and word-of-mouth. After giving informed consent and providing demographic information,
participants were taken to the experiment room, where the researcher mounted NeckFace on them, adjusting the positioning of the cameras as necessary. Following this, participants completed a \textit{calibration round}, where they were placed in front of an iPhone device and asked to copy the movements seen on a video, such as moving their head up and down or raising their eyebrows (approximately 5 minutes), complying with the initial NeckFace work \cite{neckface}.  After this, the researcher would leave the room and data collection would begin, as participants completed a \textit{stimulus round} where they watch a series of 30 videos presented in random order on a screen. Participants watched each stimulus video while a high-resolution webcam and NeckFace recorded their facial responses. Participants were not able to see their own image while the stimulus videos played. Compensation was provided at rate of USD 15/hour. After this, participants were taken back to a smaller room to fill out a final written questionnaire. The full procedure took around 30 minutes to complete. A diagram of study stages and data collected can be seen on \autoref{fig:protocol}. This data was collected under Cornell University IRB exempt protocol \#1609006604.



\subsection{System implementation}
\label{subsec:system}


In this work, both IR cameras from NeckFace are connected to a single Raspberry Pi 4B via an Arducam multi-camera adapter, different from the original NeckFace system which uses two Raspberry Pis to record the images. This allows for a lighter system that is easier to wear and integrate with other components. The two cameras capture images of the wearer's neck, chin and cheeks from both sides, which are subsequently used to estimate the full facial expressions in 3D.

The NeckFace system needs to be trained in order to provide accurate facial expression estimations. To supervise the training process, a dataset of (IR cameras image, 3D facial expression) pairs needs to be collected. Specifically, the 3D facial expressions (ground truth) were represented by 52 Blendshapes \cite{10.1145/2822013.2822042} captured with the TrueDepth camera on an iPhone 11 Pro. We also predicted three head rotation variables. We obtained the training data through the calibration round, where participants were asked to follow movements and facial expressions seen on a video (\autoref{fig:setup}).

In addition to the NeckFace device and calibration setup, a participant-facing RGB camera was used to record participants' reactions.

\subsection{Stimulus Dataset}

We selected a set of 30 stimulus videos for collecting human reactions. Following the methodology in \citet{bremers2024usingsocialcuesrecognize}, we define errors as ``actions not as intended''. For example, a robot falling while playing soccer, or a person crashing a lawnmower onto an object. We included videos of human failures (10), robot failures (10) and control videos (10) (see \autoref{fig:setup} for example). The videos' average length is $13.69\pm 7.77$ s. The full list can be consulted in the project repository \footnotemark[1].

\subsection{Participants} 
\label{subsec:parts}

A total of 30 participants completed the study. Due to technical issues or poor quality of data, 5 participants were excluded (P17 and P27 were removed because they were wearing large earrings which would shake and cause disturbances in the image quality; P27, 29 and 30 for inadequate NeckFace camera adjustment, and P13 due to erratic head motion during calibration). The remainder 25 participants took on average 33m22s$\pm$4m19s to complete the survey. Ages range from 21-78 ($37.73\pm 17.37$). 17 participants identify as female, 12 as male, and 1 as non-binary. The study included participants from 11 nationalities. Racial/ethnical distribution includes 17 Caucasian/White or Asian/White, 6 Asian/Asian American participants, 4 Hispanic/Latino and 1 African/African American/Black, 2 participants who self-described to be Indian and Indian American.

\begin{figure}
    \centering
    \includegraphics[width=1\linewidth]{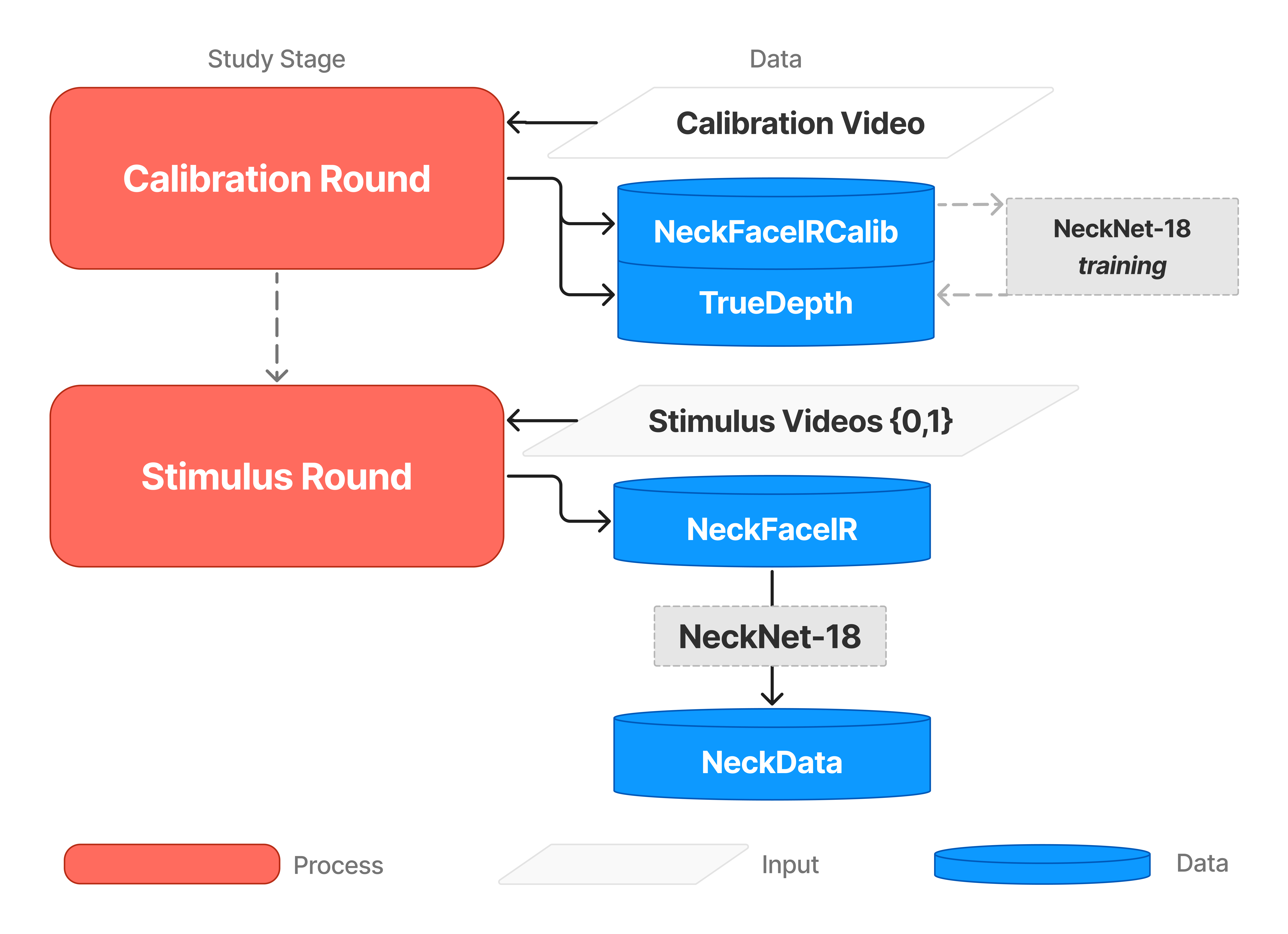}
    \caption{Study protocol and data collected. In the calibration round, NeckFace IR camera data is collected along with the Truedepth data from the Iphone (and head rotation angles). The latter serves as ground truth to train NeckNet-18. In the stimulus round, reactions to neutral (0) and error (1) videos are collected through NeckFace cameras. This dataset, NeckFaceIR, is later input to NeckNet-18, transforming the dataset into 3D facial reactions (NeckData).}
    \label{fig:protocol}
\end{figure}




\section{Model development} 
\label{sec:model}

Below, we describe model development for NeckFace 3D facial reconstruction and for error detection models. 

\subsection{3D Facial Mapping Model}
\label{subsec:neckface_model}

The 3D facial mapping model takes IR video frames captured by the NeckFace device and converts them into numerical representations of 3D facial expressions. The model was developed following the original design of NeckNet \cite{neckface}.

\begin{description}[align=left, leftmargin=0em, labelsep=0.2em, font=\textbf, itemsep=0em,parsep=0.3em]


\item[Problem Formulation and State space:] Tracking facial expressions with the wearable neckband was formulated as a regression problem: at any given time, the raw data from the wearable device is captured as two images $i_l, i_r \in \mathbb{R}_{640\times 480}$, from the left and right side camera, respectively. The model output is a feature vector $e \in \mathbb{R}_{55\times 1}$, representing the 3D facial expression and head orientation.

\item[Action Space]: The model output $e \in \mathbb{R}_{55\times 1}$ represents the 3D facial expression. This includes 52 Blendshape parameters extracted from TrueDepth camera by ARKit, representing the frontal face, and 3 angles representing the orientation (yaw, pitch, roll) of the head. The 52 Blendshapes each represent a specific facial movement (e.g., jawOpen, mouthClose, etc.), with numerical values linearly describing the extent of that movement from a scale of 0 to 1000. The ground truth of these Blendshapes was captured with the TrueDepth Camera of an iPhone 11 at 30 fps and later resampled to 12 fps to match the sampling rate of the IR cameras. A full list of the 55 features is provided as supplementary material.

\item[Dataset]: Two datasets were used to train and evaluate the facial reconstruction model. In addition to the data collected in this study, we used the original NeckFace dataset \cite{neckface} for training.

\begin{itemize}
    \item \textbf{NeckFaceOriginal}: video frames collected for \citet{neckface}. It includes images recorded from two IR cameras at 30 fps from 13 participants. This dataset was used to pre-train the model.
    \item \textbf{NeckFaceIRCalib}: video frames from NeckFace's IR cameras collected during the calibration round. Frames were recorded at 12 fps. This dataset was used to fine-tune the pre-trained model. See \autoref{fig:protocol} for schematic of how this data was used.
\end{itemize}

Data processing and augmentation were applied both during pre-training and fine-tuning. Specifically, raw images from the two cameras were first converted to grayscale and resampled to a size of 320 by 240 pixels, then tilted horizontally to generate an image of size 640x240 pixels. Data augmentation methods include: 1) 50\% chance of random scaling with a factor of 0.9 to 1.1, 2) 50\% chance of random rotation of -30 to +30 degrees (while pre-training) or -8 to +8 degrees (while fine-tuning), 3) 50\% chance of random translation of -6 to +6 pixels in both directions. These operations were applied to both sides of the images independently but with the same parameters.

\item[Evaluation Metrics]: The model was evaluated with the Mean Absolute Error (MAE) between the predicted Blendshapes and ground truth (TrueDepth). The error of the facial expressions and head orientation were calculated separately. Specifically, $\mathrm{MAE}_f = \sum_{i=1}^{52}\frac{1}{52}|e_i - \hat{e_i}|$, $\mathrm{MAE}_o = \frac{1}{3}\sum_{i=53}^{55}|e_i - \hat{e_i}|$ where $\mathrm{MAE}_f$ is the error of facial expressions tracking and $\mathrm{MAE}_o$ is the error of head orientation tracking. $e$ is the ground truth of facial expression and head orientation and $\hat{e}$ is the model prediction.

\item[Model Design]: We designed NeckNet-18, a convolutional neural network (CNN) with a ResNet-18 backbone and a fully-connected decoder (\autoref{fig:necknet}). We based this structure on the previous work \cite{neckface}, with adaptations to better fit our data and task. First, we used a lighter-weight ResNet-18 instead of the ResNet-34 model used in prior work \cite{neckface}, based on two considerations: 1) the dataset collected in this user study is smaller than that in \citet{neckface} due to smaller calibration time and a lower frame rate. Larger models would be more prone to overfitting. 2) ResNet-18 has roughly half the number of trainable parameters when compared with ResNet-34, which results in faster training and inference speed. Previous work on similar tasks also demonstrated that ResNet-18 has comparable performance to ResNet-34 with a smaller dataset \cite{eyeecho}. 

\item[Model Training]: 
NeckNet-18 is designed for within-participant generalization, using a small Calibration round to fine-tune the model for new participants. Thus, the model's generalizability relates to unseen \textit{temporal} data rather than unseen participants. 

We pre-trained the model on the \textit{NeckFaceOriginal} dataset. To fine-tune and evaluate model performance of the 3D facial mapping model, we used data collected from the calibration round, \textit{NeckFaceIRCalib}. Specifically, we first divided this dataset into 5 folds, by dividing data collected from each participant's calibration round into 5 parts in temporal order, and combining each of the 5 parts from different participants together. This means that each fold contains data from all participants, but this data does not overlap temporally. We then chose 4 of the 5 folds to use as training data to fine-tune the model, and the other fold as testing data. This process was repeated 5 times, constituting a 5-fold cross-validation where all 5 folds had appeared in the testing data once. This step also allowed us to understand the within-participant generalizability of the model. Finally, all 5 folds were combined and the model was trained again to maximize the benefit of all available training data.

During pre-training, the model was trained for 30 epochs with an initial learning rate of 0.0002. Given the differences in camera angle and image quality between the original NeckFace implementation and our re-implementation, the fine-tuning step which followed was given more weight to allow the model to better learn the distribution of the new dataset. As a result, during fine-tuning, the model was also tuned for 30 epochs but with an initial learning of 0.0001.

After training, the model was used to reconstruct the NeckFace IR data collected during the \textit{stimulus round} (participant reacted to video scenarios), which generated the dataset of 3D facial reactions used as input for the error detection model (\autoref{fig:protocol}).

\end{description}

\subsection{Error Detection Model}
\label{subsec:error detection model}

We used the output of NeckNet-18 (55 Blendshape and head motion parameters) while participants were watching stimulus videos and trained models that \textit{detect when humans are reacting to failure}. We benchmarked this performance against other conventional data inputs, and tested the models' ability to generalize to new participants or within the same participant data. \autoref{fig:protocol} illustrates how study stages and datasets integrated into our processing methodology.


\begin{description}[align=left, leftmargin=0em, labelsep=0.2em, font=\textbf, itemsep=0em,parsep=0.3em]

\item[Problem Formulation: ]

Detecting failure based on human reactions was formulated as a sequential decision-making problem: at each time step $t$, the environment is captured as a state variable $s_t \in S$ and the model output is an outcome label  $f_t \in F$, where $F$ is a discrete variable that describes the stimulus: 0 if \textit{neutral}, 1 if \textit{failure}.

\item[Action Space: ]
The reactions captured with NeckFace IR cameras and RGB video were labeled according to the type of stimulus video shown to participants. In videos where error occurs, only the moments following the exact failure moment were used as input data (and labeled \textit{1, reaction to failure}). Control videos, where there is no error, were used integrally, and labeled \textit{0, neutral state}. This reduced data imbalance across classes.

\item[Datasets: ] 

We tested 4 data input sources, to test the performance of NeckFace against other commonly used methods.

\begin{itemize}
    \item \textbf{NeckData:} this dataset was collected at 12 fps, consisting of 55 Blendshape and head motion parameters as described in Section \ref{subsec:neckface_model}, reconstructed using NeckNet-18.
    \item \textbf{OpenData:} Reactions to the stimulus videos were also recorded using a RGB camera at 30 fps. We ran OpenFace \cite{openface} on this data and obtained 49 Action Unit (AU), gaze and pose features. The dataset was undersampled to 12 fps, to match the NeckData timestamps.
    \item \textbf{NeckFaceIR:} We used video frames from NeckFace's IR cameras to test direct output classification. Cameras collected data at 12 fps and this data was resized to 224x224. Post-processing and data augmentation were performed on each frame, according to ImageNet \cite{krizhevsky2017imagenet} statistics for data normalization. 
    \item \textbf{RGBData:} We used frames from the RGB camera reaction videos, which were processed as in \textit{NeckFaceIR} and downsampled to match the 12 fps sampling and timestamps. 
\end{itemize}


We also tested normalizing the dataset and running a Principal Component (PC) Analysis for feature reduction. We kept only the PCs which explained 95\% of the variance on the dataset, resulting in 15 (resp., 33) features for the NeckData (resp. OpenData) dataset. The total number of individual datapoints per dataset can be seen in Table \ref{tab:frames}. Because all participants watched the same stimulus videos, data is balanced across participants. 

\item[Models tested: ]

To develop an error detection model from human reactions, we tested a diverse range of model architectures based on prior work in the space of robot error detection \cite{parreira2024study,spitale2024errhri2024challengemultimodal,wachowiak2024errhri}. These included Recurrent Neural Networks (RNNs) such as Long Short-Term Memory (LSTM) networks \cite{karim2018lstm}, which excel at capturing long-term dependencies; Gated Recurrent Units (GRUs), known for their efficiency in training; and Bidirectional LSTMs (BiLSTMs), which process sequences in both directions. We also explored Transformer models, leveraging their self-attention mechanism for capturing global dependencies. Additionally, we implemented MiniRocket \cite{minirocket,wachowiak2024errhri, wachowiak2024eyetrack}, a non-neural approach known for its speed and accuracy in time series classification. The gated Multi-Layer Perceptron (gMLP) \cite{liu2021payattentionmlps} was tested as an alternative to attention-based models, using spatial gating units. Lastly, we included InceptionTime and InceptionTimePlus \cite{Ismail_Fawaz_2020inception}, an ensemble of deep Convolutional Neural Network (CNN) models designed specifically for time series classification. We also tested a similar Deep Neural Network to that suggested by \citet{stiber2023erroraware}---3 hidden layers, with 64, 128 and 64 units (multi-layer DNN, ml-DNN). For the video datasets, we used ResNet34 models \cite{he2015deepresiduallearningimage}, which are 34-layer CNNs pre-trained on ImageNet \cite{krizhevsky2017imagenet}, and tested different model structures. ResNet34 is the base-model for the original version of NeckNet \cite{neckface}.




\item[Evaluation Metrics: ] The models were evaluated based on the macro averages of the following metrics: accuracy, f1-score, precision and recall. We also considered margin-of-error metrics \cite{parreira2023robot,de2012survey, spitale2024errhri2024challengemultimodal}---for a sample margin of size $k$, and for a sample $i$, the model prediction is considered correct if $y^i_{pred} \in [y^{i-k}_{pred},y^{i+k}_{pred}]$. Similar to prior works, we consider these metrics to contemplate real-life error detection scenarios, where a small tolerance in timely detection does not invalidate the system.

\begin{figure}
    \centering
    \includegraphics[width=1\linewidth]{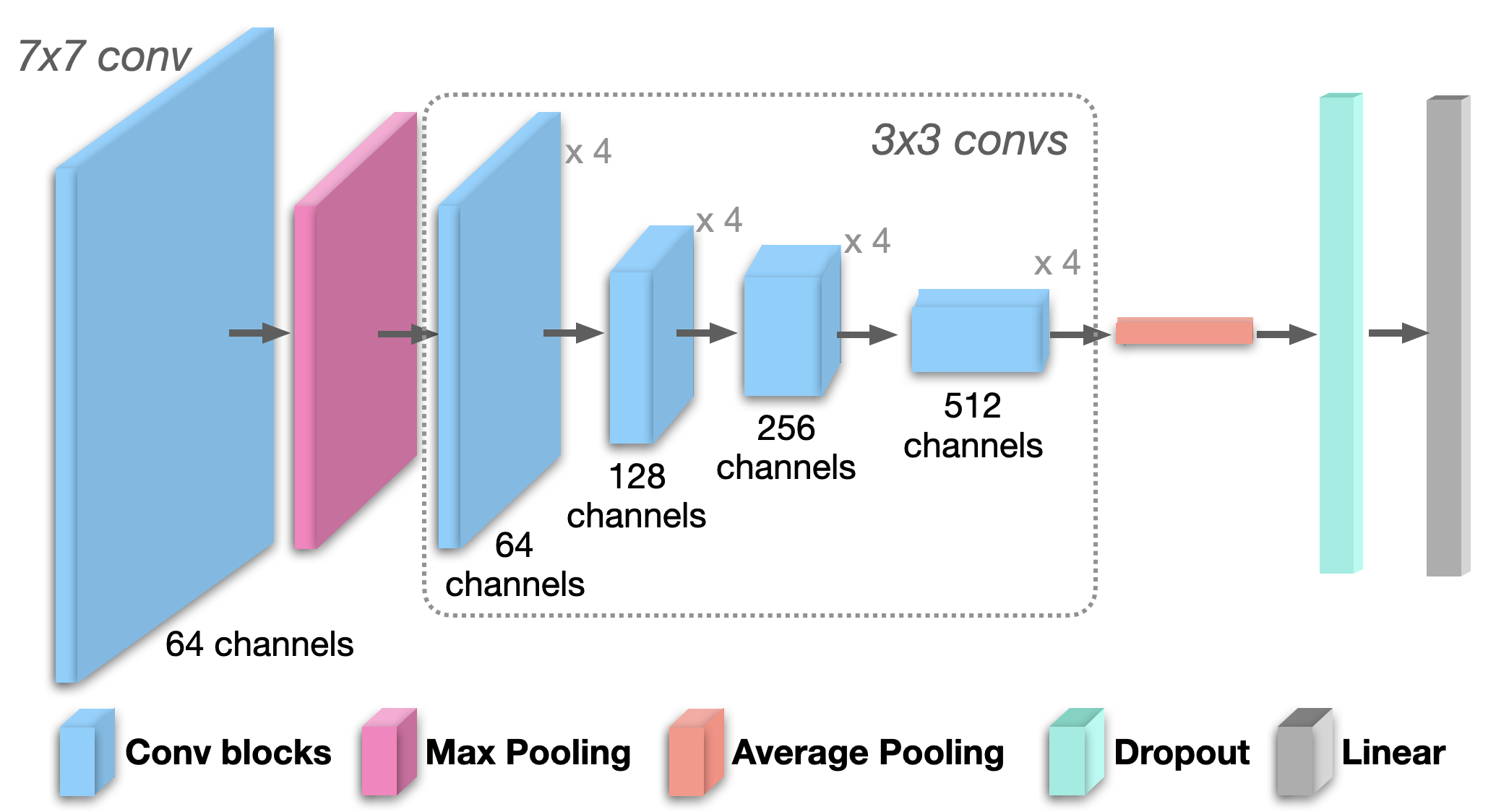}
    \caption{NeckNet-18 architecture.}
    \label{fig:necknet}
\end{figure}

\item[Model Training: ]  

We implemented model training through tsai \cite{tsai} and pytorch. For all model types and datasets, we performed hyperparameter tuning on a 70-20-10 train-val-test split, with 5 cross-validation folds with no overlapping participants, for 500 epochs. For time-series models, we implemented an interval length (i.e., number of samples in a training sample window) and stride (i.e., sliding length between sample windows) according to the methodology in \citet{wachowiak2024eyetrack}. Labels were selected according to the mode (most frequent label) of each sample window.
We picked the top-performing models based on average test accuracy across the 5 folds, and selected the best-performing epoch based on the same metric. For the Resnet34-based models (used on NeckFaceIR and RGBData datasets), best performing models were selected based on test accuracy on a single fold.

We additionally tested cross-training of the top-performing model types. That is, we used the best performing models on the \textit{NeckData} and \textit{OpenData} datasets, respectively, and fine-tuned the model's last layer to the alternative dataset, following the same process for model selection as described above.

Finally, we tested single-participant generalization, based on minimum data required for good model performance. For this, we used $n$ folds ($n =$ number of participants), and trained the models on each $i$ fold using a range between 5-45\% of randomly selected data for participant $i$. 20\% of the data was used for validation and the remaining data was used for testing the model.

\begin{figure}
    \centering
    \includegraphics[width=1\linewidth]{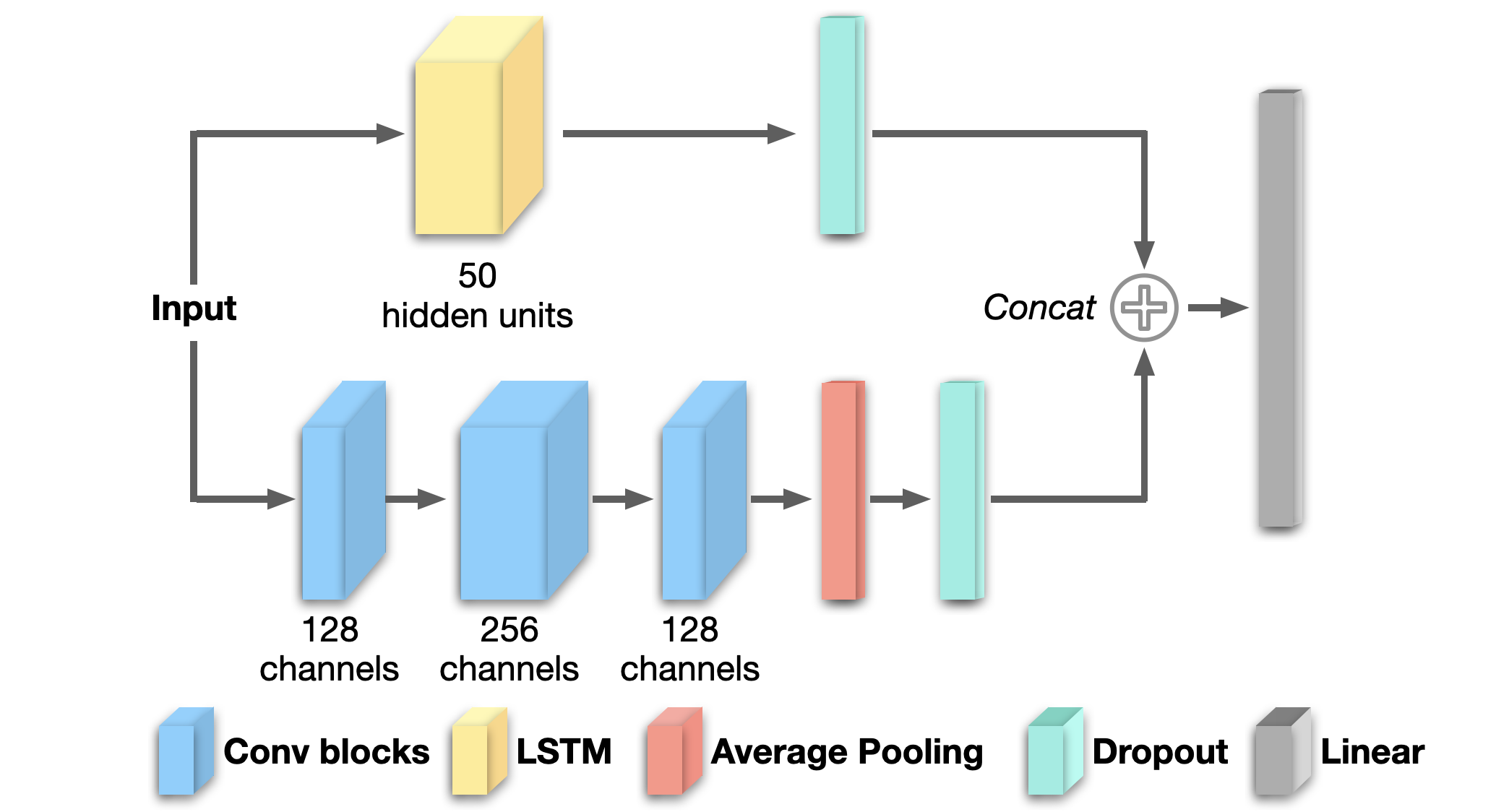}
    \caption{Best performing model for NeckData, GRU\_FCN.}
    \label{fig:error-detection-model}
\end{figure}



\item[Final Dataset for Error Detection: ]

The size of the datasets used for training are presented in \autoref{tab:frames}. NeckData is obtained from reconstructing the Neckface IR camera data from the Stimulus round using NeckNet-18. The RGB-based datasets (RGBData and OpenData) are obtained through undersampling and sample matching to NeckData.  

\begin{table}[]
\vspace{4pt}
\centering
\caption{\small Samples (frames) per dataset per type of reaction.}
\label{tab:frames}
\begin{tabular}{l|lll}
\textbf{Dataset} & \textbf{Neutral (0)} & \textbf{Error (1)} & \textbf{Total} \\ \hline
\textit{NeckData} & 51745 & 37815 & 89560 \\
\textit{OpenData} & 51580 & 37815 & 89395 \\
\textit{NeckFaceIR} & 51745 & 37815  & 89560 \\
\textit{RGBData} & 51578 & 37815 & 89393 
\end{tabular}
\end{table}

\end{description}

\section{Results} \label{sec:results}

We present the main results from our paper. Our contributions are 1) developing a 3D facial reconstruction model, NeckNet-18 and 2) benchmarking error detection models trained on NeckFace-derived data against commonly used features (e.g. OpenFace).

\subsection{3D Facial Mapping Model}
\label{subsec:results_neckface}

Following the procedures described in Section \ref{subsec:neckface_model}, we ran 5-fold cross validation (non-overlapping temporal data) to test NeckNet-18 performance for 3D facial mapping. For Blendshape parameter data (facial motion), performance was $\mathrm{MAE}_f = 34.0\pm6.9$ ($M\pm SD$), and $\mathrm{MAE}_o=7.0\pm1.3$ for head rotation angles. For reference, the original NeckNet performance \cite{neckface} was $\mathrm{MAE}_f=25.6\pm5.1$, $\mathrm{MAE}_o=3.1$.  

\subsection{Error Detection Model}

\autoref{tab:performance} presents the results from the best-performing models for each training method and dataset type. The best performing model on the NeckData dataset can be seen in \autoref{fig:error-detection-model}. \textit{Pre-trained OpenData} (resp. NeckData) refers to the model pre-trained on NeckData (resp. OpenData) and fine-tuned in OpenData (resp. NeckData). All time-series models (NeckData and OpenData datasets) performed best in the non-normalized dataset, except for pre-trained and single participant OpenData (PCA dataset). A full list of hyperparameters tested can be found in the study repository. In sum, our results for error detection from human facial cues are:
\begin{itemize}
    \item \textit{Time-series models} (trained on OpenFace features or Blendshapes): the model trained on NeckData outperforms the model trained using OpenFace features, by 5\% accuracy and 10\% F1.
    \item \textit{Pre-trained Models} (trained on one dataset and fine-tuned on the other): performance is lower than single-dataset training. Model fine-tuned on NeckData outperforms model fine-tuned on OpenData, by 5\% accuracy and 5\% F1.
    \item \textit{Single-Participant Models} (trained on subset of data from participant and tested on remaining data from that participant): for smaller training set (5\% of data used from training, 95\% for testing), NeckData model outperforms model trained using OpenFace features, by 6\% accuracy and F1. For larger training set (45\%), OpenData model slightly outperforms NeckData model, by 1\% accuracy and F1. NeckData model achieves accuracy above 84\% when trained on only 5\% of data for a single participant.
    \item \textit{Frame-Based Models} (trained on images, RGB videos from front-facing camera or videos from the IR cameras on the NeckFace device): model trained on NeckFace camera data  outperforms RGB-trained ResNet34 by 1.5\% accuracy but has lower F1 (2\%). 
    
\end{itemize}


\begin{table*}[]
\small
\centering
\caption{Model hyperparameters and test set performance. Metrics are shown in $M \pm SD$ for 5 non-overlapping participant folds, or for all participants for single-participant models ($n=25$). using data collected through the RGB camera are shown in grey rows. Percentage of participant data used for training is \textbf{5\%} for single-participant models.}
\label{tab:performance}
\begin{tabular}{llrrrr}
\hline
\multicolumn{6}{c}{\textit{Time-series Models}} \\ \hline
{\color[HTML]{000000} \textbf{Model}} & {\color[HTML]{000000} \textbf{Dataset}} & \multicolumn{1}{l}{{\color[HTML]{000000} \textbf{Accuracy}}} & \multicolumn{1}{l}{{\color[HTML]{000000} \textbf{Precision}}} & \multicolumn{1}{l}{{\color[HTML]{000000} \textbf{Recall}}} & \multicolumn{1}{l}{{\color[HTML]{000000} \textbf{F1}}} \\
\rowcolor[HTML]{EFEFEF} 
{\color[HTML]{000000} gMLP} & {\color[HTML]{000000} \textbf{OpenData}}  & {\color[HTML]{000000} $0.606 \pm 0.051$} & {\color[HTML]{000000} $0.563 \pm 0.025$} & {\color[HTML]{000000} $0.632 \pm 0.072$} & {\color[HTML]{000000} $0.535 \pm 0.015$} \\
{\color[HTML]{000000} GRU\_FCN} & {\color[HTML]{000000} \textbf{NeckData}}  & {\color[HTML]{000000} $0.658 \pm 0.061$} & {\color[HTML]{000000} $0.649 \pm 0.060$} & {\color[HTML]{000000} $0.667 \pm 0.052$} & {\color[HTML]{000000} $0.637 \pm 0.070$} \\ \hline
\multicolumn{6}{c}{\textit{Pre-trained Models}} \\ \hline
\rowcolor[HTML]{EFEFEF} 
{\color[HTML]{000000} gMLP} & {\color[HTML]{000000} \textbf{OpenData}} & {\color[HTML]{000000}  $0.569 \pm 0.040$} & {\color[HTML]{000000} \  $0.547 \pm 0.028$} & {\color[HTML]{000000}  $0.572 \pm 0.075$} & {\color[HTML]{000000} $0.538 \pm 0.019$} \\
{\color[HTML]{000000} gMLP} & {\color[HTML]{000000} \textbf{NeckData}} & {\color[HTML]{000000}  $0.619 \pm 0.057$} & {\color[HTML]{000000}  $0.597 \pm 0.042$} & {\color[HTML]{000000}  $0.619 \pm 0.057$} & {\color[HTML]{000000}  $0.589 \pm 0.047$} \\ \hline
\multicolumn{6}{c}{\textit{Single-Participant Models}} \\ \hline
\rowcolor[HTML]{EFEFEF} 
{\color[HTML]{000000} InceptionTime} & {\color[HTML]{000000} \textbf{OpenData}} & {\color[HTML]{000000} 
$0.788 \pm 0.049$} & {\color[HTML]{000000}  $0.782 \pm 0.048$} & {\color[HTML]{000000}  $0.785 \pm 0.050$} & {\color[HTML]{000000}  $0.782 \pm 0.049$} \\
\rowcolor[HTML]{EFEFEF} 
{\color[HTML]{000000} GRU\_FCN} & {\color[HTML]{000000} \textbf{NeckData}} & {\color[HTML]{000000} 
 $0.847 \pm 0.048$} & {\color[HTML]{000000}  $0.840 \pm 0.048$} & {\color[HTML]{000000}  $0.846 \pm 0.050$} & {\color[HTML]{000000}  $0.842 \pm 0.049$} \\ \hline
\multicolumn{6}{c}{\textit{Frame-Based Models}} \\ \hline

\rowcolor[HTML]{EFEFEF} 
{\color[HTML]{000000} ResNet34} & {\color[HTML]{000000} \textbf{RGB}} & {\color[HTML]{000000}  $0.515 \pm 0.025$} & {\color[HTML]{000000}  $0.505 \pm 0.020$} & {\color[HTML]{000000}  $0.505 \pm 0.018$} & {\color[HTML]{000000}  $0.493 \pm 0.023$} \\
ResNet34 & \textbf{NeckFace\_IR} & $0.530 \pm 0.031$&  $0.504 \pm 0.027$&  $0.501 \pm 0.016$&  $0.474 \pm 0.014$
\end{tabular}
\end{table*}

 \section{Discussion} \label{sec:discussion}

Motivated by human social competence, where we continuously interpret external social cues from our peers to navigate social environments, our study explored a novel approach to more contextually-aware robots through error detection by leveraging human facial reactions captured through a neck-mounted device, NeckFace. Our contributions are twofold: 1) we adapted NeckNet \cite{neckface} to a new dataset of human reactions, using a lighter model; 2) we developed models using data from the NeckFace device and NeckNet to detect errors from human reactions, finding good model performance both across and within participant data and outperforming conventional methodologies such as OpenFace.

\subsection{3D Facial Mapping Model} The performance for NeckNet-18 was lower than that of the original NeckFace work \cite{neckface}. This difference may be attributed to the following factors: 1) the original NeckFace work used more calibration data (7-minute rounds) while our calibration session had a duration of 5 minutes, of which only 4 minutes were initially used for training, 2) the original NeckFace device sampled at 30 fps while our device sampled at 12 fps. This is due to our use of a single Raspberry Pi device that records data from two cameras instead of using two Pi's. While implying a lower 3D mapping performance, these two adjustments were made to minimize the burden on participants regarding the duration of the session and device size, while preserving valid performance. According to \citet{neckface}, when MAE is under 40, the reconstructions should appear visually similar to the ground truth.

\subsection{Error Detection Model}

We further expanded on prior work on robot error detection through human social cues by extensively exploring the potential of NeckFace-generated data in reaction classification. When comparing cross-participant performance, a GRU model trained on \textit{NeckData} outperforms a gMLP model trained on \textit{OpenData}, i.e. facial action units data extracted from video frames of the participants' reactions recorded on a participant-facing camera, achieving a test accuracy of 65.8\%. While low generalizability to unseen participants has been discussed \cite{bremers2023bystander}, highlighting the need for larger datasets of human reactions to different error contexts, this performance is also higher than reported on previous works \cite{parreira2024study, spitale2024errhri2024challengemultimodal}. Notably, an experimental factor which could impact this result is different levels of participant engagement. For example, \citet{parreira2024study} collected data online through Prolific \footnote{https://www.prolific.com/}, wherein participants might be distracted or less engaged than when watching videos in-lab. Nonetheless, these results validate NeckFace as a device that effectively captures human reactions and NeckNet as a data reconstruction methodology in the context of this user study.

Interestingly, the pre-trained models (those trained on one dataset--- \textit{OpenData, NeckData}---and fine-tuned on the other) did not outperform single-dataset models. Still, we find that a model trained on OpenFace data and fine-tuned using NeckFace data outperforms a model trained on OpenFace data only. This indicates that the NeckFace approach may offer benefits even when utilizing transfer learning techniques. For frame-based models, interestingly, performance was comparable across the two datasets (video frames from an RGB participant-facing camera and from NeckFace's IR camera), with both datasets generating models which perform slightly above chance.

The most striking results come from the single-participant models. Using just 5\% of the data for training, the GRU model trained on NeckFace data achieved an accuracy of 84.7\%, surpassing the InceptionTime model trained on OpenData (78.8\%). This gap narrows when using 45\% of the data for training, suggesting that NeckFace is most useful in contexts of few calibration data, allowing for effective personalized error detection systems. Further, the GRU model used only has around 300,000 trainable parameters, which yields potential for lightweight computation and system portability.

The performance of NeckFace-based models, especially in single-participant scenarios, highlights the potential of this approach for creating more adaptable and personalized robot interaction systems. By capturing facial cues from a unique angle, NeckFace appears to provide rich, informative data that can be leveraged for error detection, namely through mapping reactions into a 3D plane rather than OpenFace's 2-dimensional processing. 
By tapping into previously underutilized data sources, we can enhance robots' ability to interpret human reactions and adjust their behavior accordingly. 

\subsection{General System Considerations}

Detecting and responding to social errors is a critical component of creating socially competent robots that can seamlessly integrate into diverse human environments. Our use of human facial reactions as a signal for error detection builds directly from HRI literature \cite{stiber2023erroraware, bremers2023bystander, parreira2024badidea}. While the technical approach leverages machine learning, the core objectives and implications of our work are deeply rooted in human-robot interaction applications. The inclusion of human and robot error videos in the stimulus dataset explores the potential of a generalized error detection system which would allow agents to contextualize failures in their environment. Future work may explore higher error granularity, e.g. through detecting robot error type \cite{2017trung}. To the best of our knowledge, no prior work has explored human reactions for multi-class error detection in the environment, which could improve robot's social and functional performance.

We acknowledge that the proposed system implies additional steps and computation, including the calibration round and the 3D reconstruction using NeckNet-18 before error detection. This work aimed to push the boundaries of input data modalities used in HRI systems by exploring the NeckFace device and its functionalities. We verify that error detection accuracy in models trained on NeckData is higher by 12-15\% than when directly using frame-based datasets. 
Similar results on frame-based models for non-overlapping participant testing had been reported \cite{bremers2023bystander, parreira2024study}. While these are non exhaustive results and there is potential for RGB-based models to detect error effectively, higher computational demands for feature extraction might be justified by better performing systems. This tradeoff warrants discussion, especially in more critical application scenarios.

\subsection{Limitations and Future Work}
\label{sec:limit}

It is important to highlight the scope and limitations of our work. 
The controlled environment of our user study may not fully represent the complexities of real-world human-robot interactions. For example, NeckFace is highly sensitive to lighting variations, which can compromise the performance of NeckNet. Further data collection in different environmental conditions, as well as additional research (e.g., in-person errors) is needed to validate the effectiveness of this approach across diverse populations and in various real-world scenarios, where noisy reactions are common.

The exclusion of participants from our dataset due to the inability to effectively transform the NeckFace camera data onto the 3D reconstruction brings about the tradeoff between larger datasets versus better-curated datasets. A longer calibration round or higher data frame rate could also provide more accurate NeckFace data reconstruction.

Finally, we note the ethical and sustainability implications of our work. While the envisioned use-case -- more harmonious human-robot interactions, where robots play a collaborative and complementary role to human activities and needs, should contribute to increased human satisfaction, fulfillment, and efficiency, there is a risk of misuse of this technology. Broader discussions ought to be held to ensure frameworks where new tools are not misappropriated and contribute to a more just and equitable society. Related to this, access to technology is also an important indicator of development and equity. The NeckFace device is relatively inexpensive and easy to replicate, and while it requires high power consumption in its current version, newer and more efficient materials should mitigate these limitations.

\section{Conclusion} \label{sec:conclusion}

This research underscores our claim that expanding human-in-the-loop robot sensing can lead to more adaptable and personalized robot interaction systems. Leveraging the social competence humans exhibit through interpreting social cues, NeckFace offers a new perspective on capturing these previously untapped data sources. The performance of NeckFace-derived models, especially in single-participant scenarios where personalization is key, reinforces its potential for creating more context-aware robots. By effectively capturing facial cues, NeckFace could enable robots to better understand and respond to human reactions in real-time. This capability is essential for the integration into dynamic human environments, pushing robots beyond simple task execution toward becoming socially aware collaborators guided by human behavior. The NeckFace form factor and classification results give confidence for further exploration into other HRI domains, where traditional methods have previously been limited.


\bibliographystyle{abbrvnat}
\balance
\bibliography{bibliography.bib}


\end{document}